\pdfoutput=1

\documentclass[]{openmoss}
\setlogos{%
    \includegraphics[height=8mm]{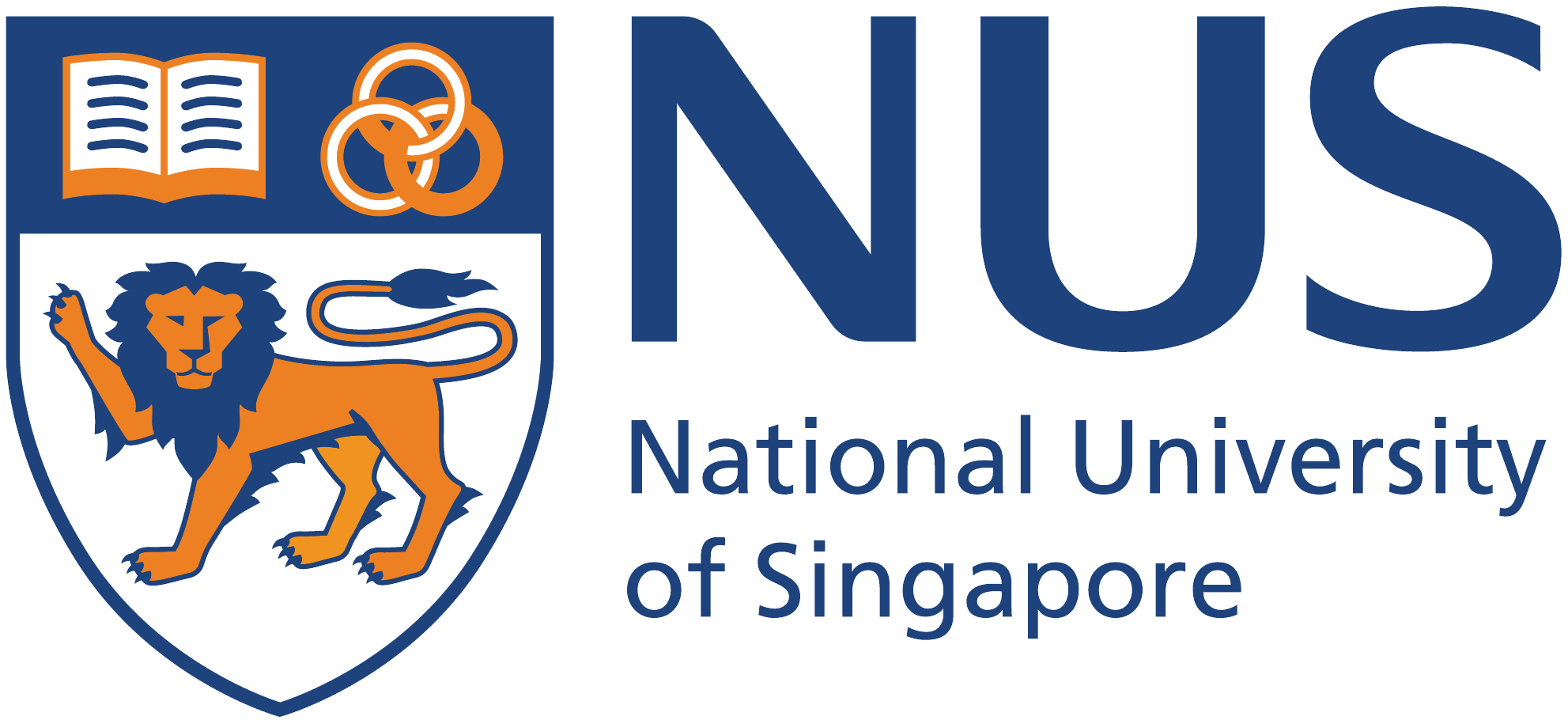}%
    \hspace{4mm}%
    \includegraphics[height=8mm]{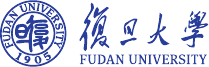}%
    \hspace{4mm}%
    \includegraphics[height=8mm]{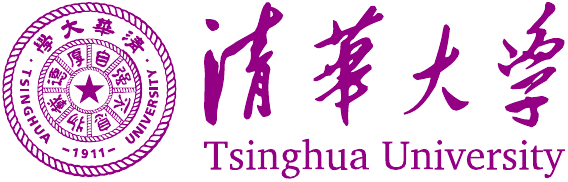}%
    \hspace{4mm}%
    \includegraphics[height=8mm]{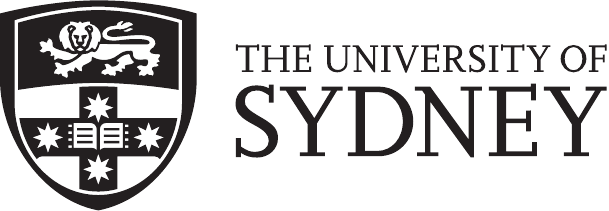}%
}
\usepackage{helvet}

\usepackage[toc,page,header]{appendix}
\usepackage{titletoc}
\usepackage{minitoc}

\usepackage{algorithm}
\usepackage{algpseudocode}

\usepackage{wasysym}
\usepackage{fontawesome}

\usepackage{enumitem}

\usepackage{listings}

\usepackage{pgfplots}
\pgfplotsset{compat=1.18}
\usetikzlibrary{arrows.meta, positioning, calc, fit, backgrounds, shapes.geometric, decorations.pathmorphing}

\usepackage{arydshln}
\usepackage{adjustbox}
\usepackage{colortbl}
\usepackage{multirow}

\usepackage{xspace}
\usepackage{nicefrac}
\usepackage{float}
\usepackage{placeins}
\usepackage{comment}
\usepackage{wrapfig}
\usepackage{pifont}
\usepackage{dsfont}

\usepackage{amsmath, amssymb}
\usepackage{amsthm}

\newcommand{\model}{\textit{World Potential Model}}
\newcommand{\wpm}{\textit{WPM}}

\usepackage{tabularx}
\usepackage{xcolor}
\definecolor{case_red}{RGB}{190,0,0}
\definecolor{case_blue}{RGB}{0,102,204}
\definecolor{case_green}{RGB}{0,153,0}

\lstdefinestyle{promptstyle}{
    basicstyle=\ttfamily\scriptsize,
    breaklines=true,
    breakatwhitespace=false,
    columns=fullflexible,
    keepspaces=true,
    showstringspaces=false,
    frame=single,
    rulecolor=\color{black!25},
    xleftmargin=0.5em,
    xrightmargin=0.5em,
    aboveskip=0.5em,
    belowskip=0.5em,
    backgroundcolor=\color{gray!10}
}

\title{World Potential Model: Pretrained World Knowledge as Progress Potentials}
\author{
    Jun Zhao\textsuperscript{1,*,$\dagger$}, Jixin Tang\textsuperscript{2,*}, Yang Shu\textsuperscript{2}, Jinyang Wu\textsuperscript{3}, Yuyang Lu\textsuperscript{2}, Jingqi Tong\textsuperscript{2},\\ Hao Xu\textsuperscript{4}, Weifeng Ge\textsuperscript{2}, Qi Zhang\textsuperscript{2}.
}
\affiliation[1]{\mbox{National University of Singapore}}
\affiliation[2]{\mbox{Fudan University}}
\affiliation[3]{\mbox{Tsinghua University}}
\affiliation[4]{\mbox{The University of Sydney}}

\contribution{* Equal contribution. $\dagger$ Corresponding author.}

\abstract{
Long-horizon language agents often receive supervision only from terminal task outcomes, leaving little signal for distinguishing productive intermediate behavior from stagnation or even regression. Rather than learning a separate value function or process reward model for every task, we ask whether pretrained models can recognize task progress from their existing world knowledge. We formalize this capability with a \model{} (\wpm{}), a goal-conditioned evaluator of task-relative realized progress in agent contexts. In \textsc{ALFWorld} and \textsc{ScienceWorld}, off-the-shelf pretrained models substantially outperform chance at recovering realized-progress structure without task-specific evaluator fine-tuning. We further anchor these progress judgments to task-specific milestones to obtain scalar \textit{world potentials}, whose temporal differences provide process-sensitive step-level credit for policy optimization. Under matched comparisons, \wpm-guided optimization improves success over outcome-only GRPO across all evaluated configurations. Together, these results provide initial evidence that pretrained world knowledge can support reusable realized-progress evaluation and provide useful supervision for long-horizon agents.
}
\contact{\email{jzhao.nlp@gmail.com}}
\coderepository{\url{https://github.com/Gabrile166/world-potential-model}}

\begin{document}
\maketitle
\vspace{-9mm}
\begin{figure}[h]
  \centering
  \includegraphics[scale=0.52]{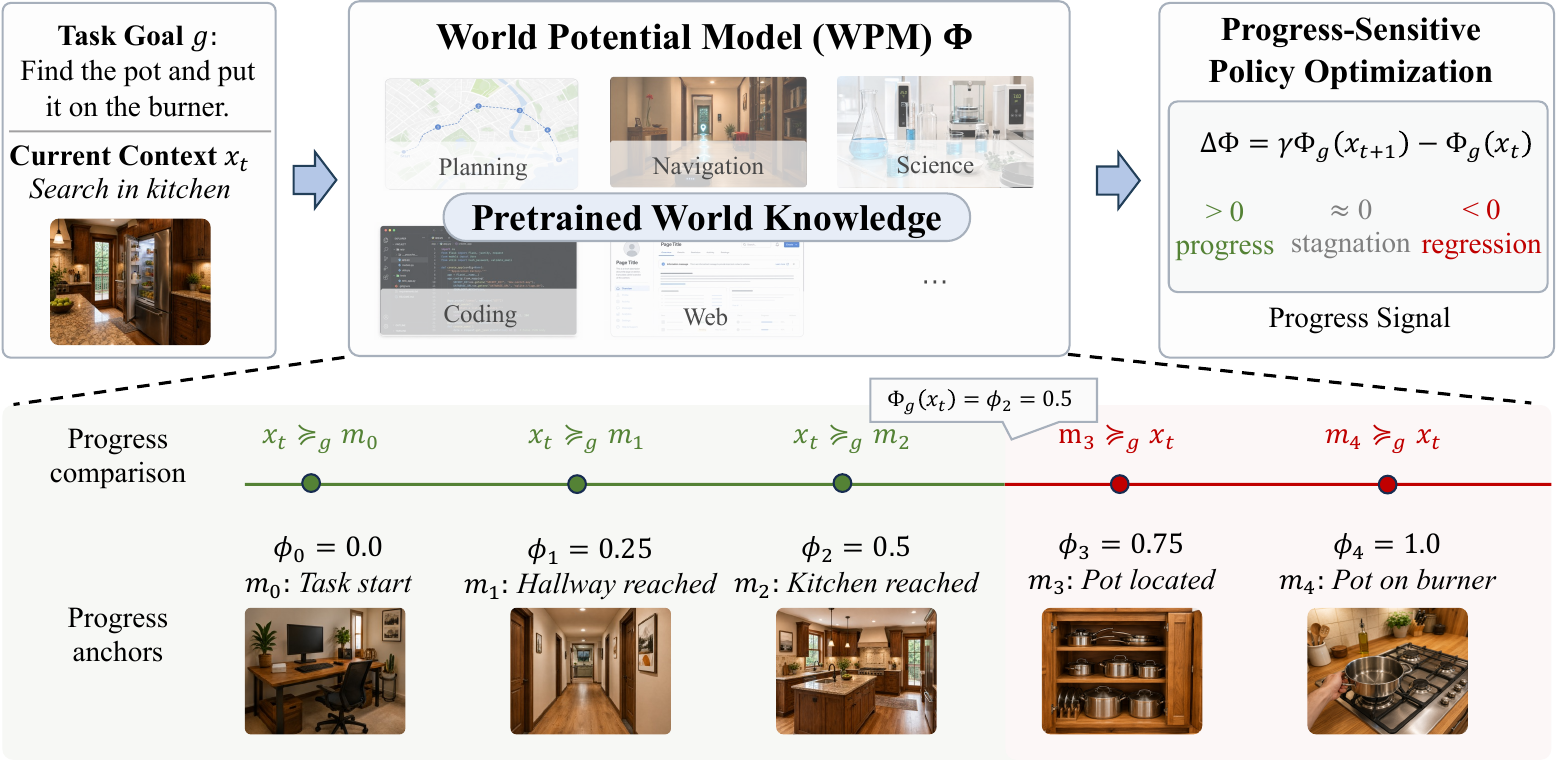}
  \caption{Overview of the \model{} (\wpm) and its downstream scalarization. A \wpm{} elicits task-relative realized-progress structure from pretrained world knowledge, with ordinal progress judgments as its primary interface. For downstream use, these judgments can be referenced to task-specific progress anchors to obtain scalar world potentials, whose temporal differences provide process-sensitive credit for policy optimization.
  }
  \label{fig: main}
\end{figure}
\section{Introduction}

Large language models (LLMs) are rapidly becoming the cognitive core of agentic systems \cite{Wang2024}, where they interpret environments \cite{pmlr-v205-huang23c}, invoke tools \cite{schick2023toolformer,Paranjape2023ARTAM}, and drive multi-step reasoning \cite{yao2023react,zhao-etal-2024-longagent,wang2024voyager} across increasingly complex tasks. As these tasks become longer-horizon, however, terminal outcomes alone provide too little structure for evaluating intermediate processes \cite{rajasekaran2026harnessdesign,young2025effectiveharnesses,zhou2026externalizationllmagentsunified,ju2026wispaperaischolarsearch,merrill2026terminalbench,li2026agencybenchbenchmarkingfrontiersautonomous,chandwani2026binarycorrectnessscalingevaluation,tan2026hindsightcreditassignmentlonghorizon}. A failed trajectory may have made substantial progress before a late mistake, while a successful trajectory may contain redundant exploration, prolonged stagnation, or temporary regression. Long-horizon agents therefore need a way to distinguish productive intermediate behavior before the final outcome is known.

Value functions and process reward models provide one route to denser supervision by estimating future return or scoring intermediate behavior \cite{10.5555/3600270.3602281,lightman2023letsverifystepstep}. These evaluators can be effective, but typically rely on task-specific annotation and training. For agents expected to operate across heterogeneous tasks and environments, repeatedly constructing a new evaluator may be expensive and difficult to scale. This raises a complementary question: \textit{must progress evaluation itself require learning a task-specific evaluator, or can it be elicited from knowledge already acquired during pretraining?}

Pretrained foundation models encounter broad descriptions of procedures, physical interactions, causal relationships, and the structure of complex tasks \citep{huang2022language,pmlr-v205-ichter23a,rocamonde2024visionlanguage,ICLR2025_48284790}. Such knowledge provides a plausible basis for recognizing not only whether an individual action is plausible, but also what has already been accomplished in an evolving agent context. For example, acquiring a required object, resolving a constraint, or completing a necessary subprocedure may represent meaningful progress, whereas undoing a satisfied condition may represent regression. This suggests that pretrained models may support a reusable capability for recognizing realized task progress without task-specific evaluator training.

We formalize this hypothesized capability through the abstraction of a \model{} (\wpm{}). Given a task goal, a \wpm{} evaluates the goal-relevant progress already realized in an agent context using pretrained world knowledge. We call the resulting task-relative progress structure \textit{world potential}. Because globally calibrated scalar progress is difficult to define across open-ended tasks, we take ordinal judgments of relative progress as the primary interface to this structure. To obtain a scalar potential for downstream use, we introduce one concrete instantiation that anchors these judgments to ordered task-specific milestones. The resulting task-local potentials can then provide step-level progress signals through their temporal differences.

We study the \wpm{} abstraction and its downstream utility in \textsc{ALFWorld} \cite{shridhar2021alfworld} and \textsc{ScienceWorld} \cite{wang-etal-2022-scienceworld}, two structured long-horizon environments in which privileged progress signals enable controlled evaluation. Across both environments, off-the-shelf pretrained models substantially outperform chance at recovering realized-progress structure without task-specific evaluator fine-tuning, providing evidence for the existence of the progress-recognition capability underlying \wpm s. We further find that, when these judgments are anchored to task-specific milestones, the resulting scalar potentials provide useful downstream supervision: under matched comparisons, \wpm-guided optimization improves success over outcome-only GRPO across all evaluated configurations.

This paper makes three contributions. First, we introduce the \model{}, a general abstraction for eliciting task-relative realized progress from pretrained world knowledge. Second, we develop a practical framework for turning realized-progress judgments into task-local scalar potentials for long-horizon agents. Third, we show how \textit{world potentials} can provide process-sensitive supervision for sparse-reward policy optimization. Together, our results provide initial evidence that realized-progress evaluation is a reusable capability of pretrained models and a useful source of supervision for long-horizon agents.

\section{Problem Setup and World Potential Models}
\label{sec:wpm_setup}
\subsection{Long-Horizon Sparse-Reward Agent Tasks}
\label{sec:task_definition}
We consider a finite-horizon episodic task specified by a goal or instruction \(g\). The environment may be partially observable. At step \(t\), let \(x_t\) denote the agent-visible interaction context, comprising the information available from the interaction history, including observations, previous actions, tool outputs, intermediate reasoning, memory, and other task-relevant information. The policy conditions on both the task goal and the current context,
\begin{equation}
    a_t \sim \pi(\cdot \mid g,x_t).
\end{equation}
After the action is executed, the environment returns new feedback or observations, which are incorporated into the interaction context to form \(x_{t+1}\). A trajectory is therefore represented as
\begin{equation}
    \tau = (x_0, a_0, x_1, a_1, \ldots, x_T).
\end{equation}
This formulation accommodates embodied, web, software, and tool-using agents whose decisions depend on an evolving interaction context rather than a compact fully observed environment state. We do not require the agent-visible context itself to be Markov.

We focus on sparse, outcome-only supervision:
\begin{equation}
r_t^{\mathrm{env}}=
\begin{cases}
0, & t<T-1,\\
R(\tau), & t=T-1,
\end{cases}
\end{equation}
where \(R(\tau)\) denotes the terminal task outcome and is binary in our experiments. Such feedback determines whether the episode ultimately succeeds, but provides little direct information about the goal-relevant progress realized at intermediate contexts.

\subsection{World Potential Models}
\label{sec:wpm}
\noindent\textbf{World potential.} For a task goal \(g\), a \model{} (\wpm) is an evaluator that uses pretrained world knowledge to assess an agent context \(x\) in terms of the goal-relevant progress already realized toward \(g\). We refer to the resulting task-relative progress structure as \textit{world potential}. The target is \textit{retrospective}: it characterizes what has already been accomplished in the current context, rather than predicting future \textit{return}. A \wpm{} need not directly produce a globally calibrated numerical progress score. Instead, we take ordinal judgments of relative realized progress as its primary interface, and introduce a task-local scalar representation only when needed for downstream use.

\noindent\textbf{Ordinal progress interface.} Given two contexts \(x_i\) and \(x_j\) from the same task instance, a \wpm{} compares their relative realized progress. We write
\begin{equation}
    x_i \succeq_g x_j,
\end{equation}
when the \wpm{} judges \(x_i\) to embody at least as much goal-relevant realized progress toward \(g\) as \(x_j\). Relative progress can be meaningful even when an absolute progress score is not. For example, after an agent acquires a necessary clue, it may be natural to judge the resulting context as representing greater realized progress than the preceding one, while assigning either context a globally meaningful score such as “0.63 complete” is considerably less well-defined. The relation is task-relative: a \wpm{} need not place unrelated tasks on a common numerical scale, but only compare realized progress within the task being evaluated. Listing~\ref{lst:pairwise_prompt} in Appendix~\ref{app:wpm_prompts} gives the pairwise prompt used in our implementation.

\subsection{Anchored Scalar World Potentials}
\label{sec:anchored_scalarization}
The ordinal \wpm{} interface in Section~\ref{sec:wpm} does not require a numerical notion of progress. For downstream applications that require a scalar signal, we introduce a task-local reference frame consisting of ordered progress anchors,
\begin{equation}
M_g=\{(m_k,\phi_k)\}_{k=0}^{K},
\end{equation}
where each \(m_k\) describes a verifiable goal-relevant accomplishment and \(\phi_k\) assigns it a task-local scalar coordinate. The anchors range from task initiation to completion, with \(m_0\) representing no meaningful progress and \(m_K\) representing task completion, and
\begin{equation}
    0=\phi_0<\phi_1<\cdots<\phi_K=1.    
\end{equation}
These coordinates need only be meaningful within the task; no numerical calibration across unrelated tasks is assumed.

Given an intermediate context \(x_t\), the \wpm{} localizes its realized progress relative to the ordered anchors. We define
\begin{equation}
    z_t=\max\{k\mid x_t\succeq_g m_k\},    
\end{equation}
where \(z_t\) denotes the most advanced anchor relative to which \(x_t\) is judged to embody at least as much goal-relevant realized progress. The corresponding scalar \textit{world potential} is
\begin{equation}
    \Phi_g(x_t)=\phi_{z_t}.
\end{equation}

This construction separates task-local calibration from progress evaluation. The anchors specify a sparse reference frame for the task, while the frozen \wpm{} supplies the evaluative mapping that places arbitrary agent contexts onto that frame using pretrained knowledge. 

The formulation itself is agnostic to how the reference frame is obtained. In our experiments, we instantiate \(M_g\) once per task from the task description and a single successful expert trajectory. The resulting anchors are then fixed, and the successful trajectory is not provided when the \wpm{} evaluates new agent contexts. Appendix~\ref{app:milestone_construction} describes this construction in detail.

\subsection{Relation to Existing Abstractions}
\model s are related to \textit{value functions}, \textit{reward models}, and \textit{world models}, but differ in what they evaluate and how that evaluation is obtained.

\noindent\textbf{Value functions.} Both \textit{value functions} and \wpm s evaluate an agent context or state, but their targets differ fundamentally. A policy-conditioned \textit{value function} \(V^\pi(x)\) is \textit{prospective}: it estimates the expected future \textit{return} obtainable from \(x\) under policy \(\pi\), and therefore generally depends on the policy being evaluated. A \wpm{} is \textit{retrospective}: it evaluates the goal-relevant progress already realized in the current context rather than predicting future \textit{return}. This progress judgment is defined independently of the policy that produced the context. Two contexts may therefore embody similar realized progress even if their future prospects under a particular policy differ substantially.

\noindent\textbf{Reward and process-reward models.} \textit{World potentials} can be converted into reward-like signals, for example through temporal differences such as \(\gamma\Phi_g(x_{t+1})-\Phi_g(x_t)\), but the \wpm{} itself is an evaluator of realized progress rather than a \textit{reward model} trained for a particular task. Conventional \textit{reward} or \textit{process-reward models} are typically fitted from task-specific labels, preferences, or rollouts. In our setting, the evaluator parameters receive no task-specific fine-tuning; instead, progress judgments are elicited from a pretrained model and, for the scalar instantiation above, referenced to task-specific milestone anchors. Thus, task-specific structure enters through the scalarization reference frame rather than through evaluator parameter training.

\noindent\textbf{World models.} Despite the shared term \textit{world}, a \wpm{} is not a model of environment dynamics. Conventional \textit{world models} predict transitions, future observations, or other consequences of actions, for example through \(p(s_{t+1}\mid s_t,a_t)\). A \wpm{} neither predicts transitions nor simulates future states. It evaluates the realized progress represented in the current context. \textit{World} instead refers to the pretrained world knowledge from which these progress judgments are elicited.

\section{Do Pretrained Models Expose World-Potential Structure?}
\label{sec:probing}
The central empirical question behind the \wpm{} abstraction is whether pretrained models can recover meaningful task-relative progress structure from agent contexts without task-specific evaluator fine-tuning. We evaluate this question at the two levels introduced in Section~\ref{sec:wpm_setup}. \textit{Pairwise progress comparison} directly tests whether pretrained models can make the relative progress judgments that constitute the \wpm's primary ordinal interface. \textit{Anchored progress localization} further tests whether these judgments can be referenced to a fixed task-specific milestone ladder to obtain the task-local localization used in the scalar construction.

\begin{table*}[t]
\centering
\small
\caption{
Accuracy (\%) on two realized-progress evaluations. \textit{Pairwise comparison} directly evaluates the ordinal progress interface of a \wpm, while \textit{Anchored localization} evaluates context localization relative to a task-specific milestone ladder. Random Guess denotes uniform selection among the valid candidate labels for each instance and is averaged over the corresponding evaluation set.
}
\label{tab:wpm_eval}
\begin{tabular}{lccccc}
\toprule
\multirow{2}{*}{\textbf{Model}} 
& \multicolumn{2}{c}{\textsc{ALFWorld}} 
& \multicolumn{2}{c}{\textsc{ScienceWorld}} 
& \multirow{2}{*}{\textbf{Avg.}} \\
\cmidrule(lr){2-3} \cmidrule(lr){4-5}
& \texttt{Pairwise} & \texttt{Anchored} & \texttt{Pairwise} & \texttt{Anchored} & \\
\midrule
Random Guess & 50.00 & 20.00 & 50.00 & 15.27 & 33.82 \\ 
MiMo-VL-7B \cite{xiaomimimo2025mimovl7brl2508} & 92.67 & 58.00 & 97.33 & 63.33 & 77.83 \\
Qwen2.5-VL-7B \cite{qwen2025qwen25vl7binstruct} & 80.00 & 60.00 & 81.33 & 48.00 & 67.33 \\
Qwen2.5-32B \cite{qwen2024qwen2532binstructawq} & \textbf{95.33} & 80.67 & 96.67 & 76.00 & 87.17 \\
Qwen2.5-VL-72B \cite{qwen2025qwen25vl72binstructawq} & 93.33 & 82.00 & 94.67 & 81.33 & 87.83 \\
Qwen3-30B-A3B \cite{qwen2025qwen330ba3binstruct2507fp8} & 91.33 & 86.67 & 97.33 & 80.00 & 88.83 \\
Qwen3-VL-30B-A3B \cite{qwen2025qwen3vl30ba3binstruct} & 90.00 & \textbf{88.00} & \textbf{98.00} & \textbf{85.33} & \textbf{90.33} \\
Qwen3-VL-32B \cite{qwen2025qwen3vl32bthinkingfp8}& 92.67 & 86.67 & 94.67 & 70.00 & 86.00 \\
\bottomrule
\end{tabular}
\end{table*}

\subsection{Evaluation Tasks}
\noindent\textbf{Pairwise progress comparison.}
Given two agent contexts \(x_i\) and \(x_j\) associated with the same task goal \(g\), the model predicts which context embodies greater goal-relevant progress already realized toward task completion. The two contexts may arise from different interaction trajectories and need not share a common prefix. The task therefore requires comparing the realized progress represented in the contexts rather than relying on their temporal positions within a trajectory. No milestone ladder is provided, so this evaluation directly probes the ordinal progress interface defined in Section~\ref{sec:wpm}.

\noindent\textbf{Anchored progress localization.}
Given an agent context \(x_t\) together with its fixed task-specific milestone ladder \(M_g\), the model predicts the anchor index \(z_t\) corresponding to the context's realized-progress level. The number of anchors is determined during task-specific milestone construction and may vary across tasks. The corresponding scalar world potential is then given by the fixed anchor coordinate \(\phi_{z_t}\). This evaluation therefore tests the anchored scalar representation introduced in Section~\ref{sec:anchored_scalarization} rather than requiring the model to assign an unconstrained numerical progress score directly to the context.

Both tasks are formulated as categorical prediction problems. The exact evaluation prompts are provided in Appendix~\ref{app:wpm_prompts}.

\subsection{Experimental Setting}
\noindent\textbf{Benchmarks and labels.}
We instantiate both evaluation tasks in \textsc{ALFWorld} \cite{shridhar2021alfworld} and \textsc{ScienceWorld} \cite{wang-etal-2022-scienceworld}, two long-horizon interactive environments in which successful task completion requires multiple stages of meaningful intermediate progress. 
Both environments provide privileged signals for constructing environment-grounded progress labels. In \textsc{ALFWorld}, we use normalized remaining expert-plan length as a planner-based \textit{proxy}: although formally a cost-to-go measure, manual inspection indicates that its ordering tracks relative realized progress, and we filter near-tied states. In \textsc{ScienceWorld}, the native task score directly awards partial credit for satisfied goals and subgoals. These signals are used only for evaluation-label construction and are never provided to the models; Appendix~\ref{app:capability_labels} gives full details.


\noindent\textbf{Models and evaluation protocol.}
We evaluate seven language and vision-language models from three model families: MiMo, Qwen2.5, and Qwen3. All models are tested under the same prompting protocol and without task-specific evaluator fine-tuning. We measure accuracy against environment-grounded labels. For an evaluation instance with \(C_i\) candidate labels, uniform random guessing has expected accuracy \(1/C_i\). Pairwise comparison therefore has a fixed chance accuracy of \(50\%\), whereas anchored progress localization has a task-dependent chance accuracy because the number of milestone anchors varies across tasks. The mean random-guess accuracies are \(20.00\%\) in \textsc{ALFWorld} and \(15.27\%\) in \textsc{ScienceWorld}.

\subsection{Results}

Table~\ref{tab:wpm_eval} reports performance on \textit{pairwise progress comparison} and \textit{anchored progress localization}.

\noindent\textbf{Relative realized progress is strongly recoverable in these settings.}
All evaluated models substantially outperform the \(50\%\) random baseline on pairwise progress comparison. Averaged across models and environments, pairwise accuracy reaches \(92.52\%\), with the best results reaching \(95.33\%\) on \textsc{ALFWorld} and \(98.00\%\) on \textsc{ScienceWorld}. These results provide direct evidence that pretrained models can often determine which of two partially completed agent contexts embodies greater goal-relevant realized progress, even when the contexts arise from different interaction paths. The result is consistent across all three evaluated model families, both language-only and vision-language models, and both environments, suggesting that it is not specific to a single model or benchmark.

\noindent\textbf{Anchored progress localization is harder but remains informative.}
Average localization accuracy across models and environments reaches \(74.71\%\), with all evaluated models exceeding the corresponding random baselines of \(20.00\%\) in \textsc{ALFWorld} and \(15.27\%\) in \textsc{ScienceWorld}. The best results reach \(88.00\%\) and \(85.33\%\), respectively. Although localization accuracy is lower than pairwise accuracy, this gap is consistent with the different structure of the two evaluations: pairwise comparison requires resolving a single relative-progress relation, whereas anchored localization requires placing a context among multiple ordered, task-dependent progress anchors. These results show that pretrained models can meaningfully localize agent contexts relative to fixed task-specific milestone ladders, providing the localization step required by the scalar construction in Section~\ref{sec:anchored_scalarization}.

\noindent\textbf{Takeaway.}
The pairwise results support the core premise of the \wpm{} abstraction: in two controlled long-horizon environments, off-the-shelf pretrained models recover substantial task-relative ordinal realized-progress structure without task-specific evaluator fine-tuning. Anchored localization is less accurate but remains substantially above chance, showing that these judgments can also be referenced to task-specific progress anchors for the scalar construction introduced in Section~\ref{sec:anchored_scalarization}. We next ask whether world potentials constructed from these imperfect judgments provide useful downstream supervision.

\section{Progress-Sensitive Policy Optimization with World Potentials}
\label{sec:optimization_objective}
The anchored construction in Section~\ref{sec:anchored_scalarization} provides a task-local scalar potential
\(\Phi_g(x)\) over agent contexts. We now consider one downstream use of this
signal: converting temporal changes in world potential into process-sensitive
step-level credit for critic-free policy optimization. Section~5 uses this
construction to test whether imperfect world-potential estimates provide useful
supervision beyond sparse terminal outcomes.

\subsection{World-Potential Transition Signal}
\label{sec:transition_signal}
Under the sparse-reward setting of Section~\ref{sec:task_definition}, the environment reward is zero at intermediate steps and is determined by the task outcome on the terminal transition. We augment this sparse signal with the temporal difference in \textit{world potential}:
\begin{equation}
    r_t^{\mathrm{WP}}
    =
    r_t^{\mathrm{env}}
    +
    \gamma \Phi_g(x_{t+1})
    -
    \Phi_g(x_t),
    \label{eq:wp_transition}
\end{equation}
where \(\gamma \in (0,1]\) is the base temporal discount. A transition to a context with higher \textit{world potential} receives a positive potential contribution, whereas a transition to a lower-potential context receives a negative contribution. When the potential remains unchanged (i.e., \(\Phi_g(x_{t+1})=\Phi_g(x_t)>0\)) and \(\gamma<1\), the potential contribution becomes \(-(1-\gamma)\Phi_g(x_t)<0\), yielding a small negative signal and mildly discouraging interaction that fails to advance the task.

\subsection{Process-Sensitive Returns}
\label{sec:process_sensitive_returns}

We next aggregate the transition signals in
Eq.~\ref{eq:wp_transition} into step-level returns. To control how
future intermediate contexts contribute to the credit assigned at
step \(t\), we introduce a process-sensitivity parameter
\(\rho\in[0,1]\) and define the effective return discount as
\begin{equation}
    \beta = \gamma\rho.
    \label{eq:effective_discount}
\end{equation}
The resulting world-potential return from step \(t\) is
\begin{equation}
    G_t^{\mathrm{WP},\rho}
    =
    \sum_{\ell=0}^{T-t-1}
    \beta^\ell r_{t+\ell}^{\mathrm{WP}}
    =
    \sum_{\ell=0}^{T-t-1}
    (\gamma\rho)^\ell r_{t+\ell}^{\mathrm{WP}}.
    \label{eq:wp_return}
\end{equation}
Let \(H=T-t\). Substituting Eq.~\ref{eq:wp_transition} and
collecting the potential terms gives
\begin{align}
G_t^{\mathrm{WP},\rho}
={}&
\underbrace{
\sum_{\ell=0}^{H-1}
\beta^\ell r_{t+\ell}^{\mathrm{env}}
}_{\text{environment return}}
-
\underbrace{
\Phi_g(x_t)
}_{\text{current-context potential}}
+
\underbrace{
\gamma(1-\rho)
\sum_{k=1}^{H-1}
\beta^{k-1}\Phi_g(x_{t+k})
}_{\text{intermediate world potentials}}
+
\underbrace{
\gamma\beta^{H-1}\Phi_g(x_T)
}_{\text{terminal-context potential}}.
\label{eq:wp_return_decomposition}
\end{align}
For \(0<\rho<1\), each future intermediate context contributes explicitly to the return with weight
\begin{equation}
    w_k(\rho)
    =
    \gamma(1-\rho)(\gamma\rho)^{k-1}
    =
    \gamma^k(1-\rho)\rho^{k-1},
    \label{eq:intermediate_weight}
\end{equation}
where $1\leq k < H$ and adjacent weights satisfy
\begin{equation}
    \frac{w_{k+1}(\rho)}{w_k(\rho)}
    =
    \gamma\rho
    < 1,
\end{equation}
so the contribution of intermediate \textit{world potentials} decays geometrically with temporal distance. Smaller values of \(\rho\) concentrate this contribution more strongly on nearer-term contexts, whereas larger values extend credit over a longer portion of the remaining trajectory. The return can therefore remain sensitive to when progress is realized and how it evolves over subsequent intermediate contexts.

The boundary cases further clarify the role of \(\rho\). At
\(\rho=0\), we have \(\beta=0\), and the return reduces to the
one-step transition signal,
\begin{equation}
    G_t^{\mathrm{WP},0}
    =
    r_t^{\mathrm{WP}}.
\end{equation}
At \(\rho=1\), by contrast, \(\beta=\gamma\) and the coefficient on
every intermediate potential in
Eq.~\ref{eq:wp_return_decomposition} becomes zero. In this case,
using the same discount \(\gamma\) for both the potential difference
and its temporal accumulation causes the intermediate potential terms
to cancel telescopically, leaving
\begin{equation}
    G_t^{\mathrm{WP},1}
    =
    \sum_{\ell=0}^{H-1}
    \gamma^\ell r_{t+\ell}^{\mathrm{env}}
    -
    \Phi_g(x_t)
    +
    \gamma^H\Phi_g(x_T).
    \label{eq:wp_return_telescoping}
\end{equation}
Thus, for \(0<\rho<1\), intermediate \textit{world potentials} remain explicitly represented in the return rather than canceling out. The resulting return is therefore intentionally process-sensitive: it uses intermediate realized-progress estimates to distinguish trajectories that may share the same terminal outcome but differ in how progress unfolds along the way.

\subsection{Step-Level Group-Relative Advantages}

We retain the group-relative standardization principle of GRPO, but apply it to step-level world-potential returns rather than trajectory-level outcome rewards. Consider a rollout group of \(N\) trajectories sampled for the same task \(g\). For every valid environment step, we compute the world-potential return \(G^{\mathrm{WP},\rho}_{i,t}\) and standardize these returns across all valid environment steps from the \(N\) trajectories in the task group:
\begin{equation}
    \hat{A}_{i,t}
    =
    \frac{
        G^{\mathrm{WP},\rho}_{i,t}
        -
        \mu^{\mathrm{step}}_g
    }{
        \sigma^{\mathrm{step}}_g
        +
        \epsilon_{\mathrm{norm}}
    },
\end{equation}
where \(\mu^{\mathrm{step}}_g\) and \(\sigma^{\mathrm{step}}_g\) denote the mean and standard deviation over the corresponding set of valid step-level returns.

This changes the granularity of credit assignment relative to our outcome-only GRPO baseline. In that baseline, each trajectory receives a single group-relative advantage derived from its terminal outcome, which is shared by all actions in the trajectory. WPM-guided optimization instead assigns a distinct advantage to each valid environment step according to its process-sensitive return. Actions followed by different patterns of progress, stagnation, or regression can therefore receive different optimization weights even within the same trajectory.


\section{A Utility Case Study in Sparse-Reward Policy Optimization}
\label{sec:utility}

Section~\ref{sec:probing} shows that pretrained models can recover meaningful realized-progress structure from agent contexts, while Section~\ref{sec:optimization_objective} converts anchored \textit{world potentials} into process-sensitive step-level credit. We now ask whether this credit provides useful supervision for policy learning under sparse terminal rewards. We compare \wpm-guided optimization with outcome-only GRPO under matched actor-training configurations, using task success as the downstream evaluation criterion. This experiment is intended as a controlled test of the utility of \wpm-derived process-sensitive credit rather than as a comprehensive comparison of reinforcement learning algorithms.

\subsection{Experimental Setup}

\noindent\textbf{Environments and data splits.}
We evaluate on the same two long-horizon interactive environments used in Section~\ref{sec:probing}: \textsc{ALFWorld} and \textsc{ScienceWorld}. For \textsc{ALFWorld}, policies are trained on the official training split and evaluated on \texttt{valid\_unseen}. For \textsc{ScienceWorld}, policies are trained on the official training variations and evaluated on the corresponding test variations. Both environments provide sparse terminal supervision.

\noindent\textbf{Policies and controlled comparison.}
We optimize Qwen2.5-Instruct models at 3B and 7B scale and Qwen3 at 4B scale. For reference, we also report the corresponding vanilla checkpoints. Our primary comparison is between outcome-only GRPO and \wpm-guided optimization. For each actor model, the two optimization methods use matched task sampling, rollout construction, and actor-update configurations; they differ in the learning signal and resulting credit construction. Outcome-only GRPO derives trajectory-level advantages from terminal task outcomes, whereas \wpm-guided optimization uses the process-sensitive step-level returns defined in Section~\ref{sec:optimization_objective}. \textit{World potentials} are produced by a frozen Qwen3-VL-32B evaluator using the fixed task-specific milestone ladders described in Section~\ref{sec:anchored_scalarization}. The evaluator parameters remain fixed throughout actor training. For each actor–environment configuration, we run both methods with the same three random seeds and report mean performance across the three runs.

\noindent\textbf{Evaluation metric.}
We report task success rate, defined as the percentage of evaluation episodes
whose final environment state satisfies the task goal. Full details of task
sampling, rollout construction, optimization hyperparameters, evaluator settings,
and metric computation are provided in Appendix~\ref{app:policy_optimization}.

\begin{table}[t]
    \centering
    \caption{
    Task success rate (\%) for vanilla policies, outcome-only GRPO, and \wpm-guided optimization in \textsc{ALFWorld} and \textsc{ScienceWorld}. Results for GRPO and \wpm{} are averaged over three independent random seeds.
    }
    \label{tab:model_results}
        \begin{tabular}{c c c c}
            \toprule
            \textbf{Model}
            & \textbf{Method}
            & \shortstack{\textsc{Alf}\\\textsc{World}}
            & \shortstack{\textsc{Science}\\\textsc{World}} \\
            \midrule
            \multirow{3}{*}{Qwen2.5‑3B}
            & Vanilla & 10.93 & 8.59 \\
            & GRPO & 53.90 & 51.56 \\
            & WPM  & \textbf{60.50} & \textbf{55.39} \\
            \midrule
            \multirow{3}{*}{Qwen2.5‑7B}
            & Vanilla & 15.62 & 13.28 \\
            & GRPO & 68.70 & 66.40 \\
            & WPM  & \textbf{72.10} & \textbf{74.73} \\
            \midrule
            \multirow{3}{*}{Qwen3‑4B}
            & Vanilla & 22.65 & 19.53 \\
            & GRPO & 48.43 & 42.96 \\
            & WPM  & \textbf{53.64} & \textbf{46.09} \\
            \bottomrule
        \end{tabular}
\end{table}

\subsection{Results}

\noindent\textbf{\wpm-guided optimization consistently improves task success.} As shown in Table~\ref{tab:model_results}, \wpm-guided optimization outperforms the matched outcome-only GRPO baseline in all six actor--environment configurations. The improvements range from \(3.13\) to \(8.33\) percentage points, with an average gain of \(5.08\) points. The largest improvement occurs for Qwen2.5-7B on \textsc{ScienceWorld}, where success increases from \(66.40\%\) to \(74.73\%\). Improvements are observed for all three actor models in both environments.

These results provide evidence that the realized-progress information recovered by the \wpm{} can be converted into useful supervision for sparse-reward policy optimization. In particular, under the matched training configurations studied here, the process-sensitive credit construction of Section~\ref{sec:optimization_objective} provides more useful learning signal than assigning credit solely from terminal trajectory outcomes. The experiment therefore establishes a downstream utility result for \textit{world potentials} in the evaluated settings, without requiring the \wpm{} evaluator itself to be trained on task-specific progress labels.

\section{Related Work}

\noindent\textbf{Process supervision and progress-based credit.} Process supervision provides intermediate learning signals beyond terminal correctness \cite{lightman2023letsverifystepstep}. Recent work extends this idea to interactive agents through learned process reward models, rollout-derived credit, and subgoal- or progress-based supervision \cite{tan2026hindsightcreditassignmentlonghorizon,choudhury2025processrewardmodelsllm,xi2025agentprmprocessrewardmodels,zhang2025progrmbuildbettergui,auzina2026intrinsic,wang2026subgoaldrivenframeworkimprovinglonghorizon}. These approaches reinforce the importance of intermediate credit in long-horizon tasks, but obtain such credit from different sources, including policy rollouts, learned evaluators, intrinsic beliefs, or explicit subgoal structure. Our focus is complementary: we ask whether a frozen pretrained model already exposes a task-relative notion of realized progress that can be elicited without task-specific evaluator fine-tuning. A \wpm{} therefore treats realized-progress evaluation as the object of study itself, with ordinal comparison as its primary interface, rather than defining progress through predicted future return.

\noindent\textbf{Foundation models for reward and progress construction.} Pretrained language and vision-language models have been used to generate reward functions, provide reward judgments, and construct task-progress signals \cite{xie2024textreward,ma2024eureka,rocamonde2024visionlanguage,pmlr-v235-wang24bn,sarukkai2025automated,zhang2025progress,müller2026automatingpotentialbasedrewardshaping}. These methods share with \wpm s the premise that pretrained models contain task knowledge that can support downstream learning. Our formulation isolates a different interface to that knowledge: the pretrained evaluator makes ordinal judgments about progress already realized in an agent context, which can then be referenced to task-specific anchors when a scalar potential is needed. This separates the general progress-evaluation capability from the particular scalarization and optimization scheme used downstream. Our policy-optimization construction is also related to potential-based reward shaping \cite{645528.657613}; however, for \(0<\rho<1\), intermediate world potentials remain explicitly represented in the process-sensitive return, so the resulting objective is not the classical policy-invariant shaping construction.

\section{Conclusion}

We introduced the \model{} (\wpm{}), an abstraction for eliciting task-relative realized progress from pretrained world knowledge. In \textsc{ALFWorld} and \textsc{ScienceWorld}, off-the-shelf pretrained models recover substantial realized-progress structure without task-specific evaluator fine-tuning. We further showed that these judgments can be anchored into scalar \textit{world potentials} and converted into process-sensitive step-level credit. Under matched actor-training configurations, \wpm-guided optimization improves task success over outcome-only GRPO across all evaluated settings.

Together, these results provide initial evidence that realized-progress evaluation  can serve as a reusable interface between pretrained world knowledge and long-horizon agent learning. Evaluating \textit{world potentials} across broader tasks and agent settings remains an important direction.

\bibliographystyle{unsrtnat}
\bibliography{custom}

\newpage
\beginappendix
\section{World Potential Model Instantiation and Capability Evaluation Details}
\label{app:wpm}
This appendix provides implementation details for the \model{} (\wpm{}) instantiation and the capability evaluations in Section~\ref{sec:probing}. We first present the prompt interfaces used to elicit realized-progress judgments, then describe the construction of task-specific milestone ladders, and finally detail the environment-grounded labels used for capability evaluation.

\subsection{WPM Progress-Judgment Interfaces}
\label{app:wpm_prompts}
As described in Section~\ref{sec:wpm_setup}, we access world potential primarily through ordinal judgments rather than asking the model to directly produce a globally calibrated scalar score. Our implementation uses two prompt interfaces: pairwise progress comparison between agent contexts and anchored progress localization of a single context.

\paragraph{Pairwise progress comparison.}
Given two agent contexts associated with the same task goal, the \wpm{} identifies which context embodies greater goal-relevant realized progress toward task completion. Listing~\ref{lst:pairwise_prompt} shows the prompt used for this comparison.

\begin{lstlisting}[
    style=promptstyle,
    %float=*t,
    caption={Prompt for comparing the progress of two agent contexts.},
    label={lst:pairwise_prompt}
]
You are evaluating the progress of two agent contexts on the same task.

## Task Description
{sample.task_description}

## Context A
{context_a}

## Context B
{context_b}

## Question
Which context has made MORE progress toward completing the task?

Provide your reasoning first, then give your final answer.

Output format:
Reasoning: <your analysis>
Answer: <A or B>
\end{lstlisting}

\paragraph{Anchored progress localization.}
For downstream scalarization, we additionally provide the \wpm{} with the fixed task-specific milestone ladder and ask it to localize the current context relative to the ordered progress anchors. The output is the anchor index corresponding to the context's realized-progress level, whose associated coordinate determines the scalar \textit{world potential}. Listing~\ref{lst:localization_prompt} shows the prompt used in our implementation.

\begin{lstlisting}[
    style=promptstyle,
    %float=*t,
    caption={Prompt for anchored progress localization.},
    label={lst:localization_prompt}]
You are evaluating the realized progress of an embodied agent on a task.

## Task
{sample.task_description}

## Ordered Progress Anchors
{generated_formatted_milestones}

The anchors represent increasing levels of realized progress toward task completion.

## Current Context
{formatted_context}

## Question
Which anchor best represents the realized-progress level of the current context?

Do not assign progress that has not yet been realized.

Provide your reasoning first, then give your final answer.

Output format:
Reasoning: <your analysis>
Answer: <one anchor id, e.g. {answer_options}>
\end{lstlisting}

Pairwise comparison directly probes the ordinal progress relation underlying a \wpm{}, while anchored localization implements the task-local scalarization used in our experiments. In Section~\ref{sec:probing}, both outputs are evaluated against independently constructed environment-grounded labels described in Appendix~\ref{app:capability_labels}.

\subsection{Task-Specific Anchors Construction}
\label{app:milestone_construction}
The scalar construction in Section~\ref{sec:anchored_scalarization} requires a task-local set of ordered progress anchors \(M_g\). The WPM formulation does not require a particular source for these anchors; here we describe the construction used in our experiments.

For each task instance, we provide a pretrained model with the task description and one successful expert trajectory. The purpose of the trajectory is to expose a valid realization of the task from which the model can identify salient intermediate accomplishments. It is used only to construct the reference frame: it does not train or fine-tune the \wpm{} evaluator, and it is not provided when the \wpm{} subsequently evaluates agent-generated contexts.

Importantly, the generated anchors describe task states rather than steps of the demonstration. We instruct the model to identify verifiable accomplishments that characterize increasing levels of goal-relevant realized progress, rather than reproducing the demonstrated action sequence. Consequently, the resulting ladder provides a sparse set of reference points rather than progress labels for the arbitrary contexts that the policy may later visit. Localizing such contexts relative to these anchors remains the responsibility of the frozen \wpm{}.

Concretely, the model generates between 4 and 10 intermediate anchors according to task complexity. Each anchor must (i) describe a salient accomplishment relevant to task completion, (ii) be state-based rather than tied to a particular action or trajectory position, and (iii) be verifiable from the agent-visible context. The model also assigns each anchor a scalar coordinate satisfying

\[
0=\phi_0<\phi_1<\cdots<\phi_K=1,
\]

where \(m_0\) denotes no meaningful task progress and \(m_K\) denotes task completion. Coordinates are intended to reflect the relative amount of goal-relevant progress represented by each accomplishment rather than being determined solely by its ordinal index.

The resulting anchor set is generated once and fixed throughout capability evaluation and policy optimization. Thus, this instantiation uses a single successful demonstration to establish task-local calibration, while relying on pretrained knowledge to interpret previously unseen and potentially off-demonstration agent contexts relative to that calibration.

Listing~\ref{lst:milestone_generation} gives the prompt used to construct the anchor set, and Listing~\ref{lst:milestone_example} shows an example.

\begin{lstlisting}[
    style=promptstyle,
    %float=*t,
    caption={Prompt template for constructing a task-specific milestone ladder.},
    label={lst:milestone_generation}]
You are a task decomposition expert.

## Task Description
{task_description}

## Expert Successful Trajectory
{expert_trajectory}

## Instructions
Analyze the task and successful expert trajectory and construct an ordered set of milestones representing increasing levels of realized progress from task initiation to completion.

Requirements:
1. Generate between 4 and 10 intermediate milestones depending on task complexity: simpler tasks need fewer milestones, while complex multi-stage tasks need more.
2. Each milestone should describe a salient accomplishment or state relevant to completing the task.
3. Criteria must be state-based rather than step-based. Do not refer to step numbers or positions in the expert trajectory.
4. Milestones should be verifiable from the agent context and ordered according to the logical progression of the task.
5. Assign each milestone a progress coordinate between 0 and 1 that reflects how much goal-relevant progress the milestone represents. Coordinates must increase strictly with the milestone order; do not assign them solely from the milestone index. Use 0 for the initial anchor and 1 for task completion.

Output format (strict JSON):
{
  "milestones": [
    {
      "id": "M0",
      "name": "Not started",
      "criteria": "No meaningful task progress has been made.",
      "progress": 0.0
    },
    {
      "id": "M1",
      "name": "Milestone name",
      "criteria": "Verifiable state-based accomplishment",
      "progress": <value between 0 and 1>
    },
    ...
    {
      "id": "M<K>",
      "name": "Task complete",
      "criteria": "The task objective has been completed.",
      "progress": 1.0
    }
  ],
  "reasoning": "Brief explanation of the milestone decomposition
                and progress coordinates."
}
\end{lstlisting}

\begin{lstlisting}[
    style=promptstyle,
    %float=*t,
    caption={Example of a generated task-specific milestone ladder.},
    label={lst:milestone_example}]
{
  "milestones": [
    {
      "id": "M0",
      "name": "Not started",
      "criteria": "No meaningful task progress has been made.",
      "progress": 0.0
    },
    {
      "id": "M1",
      "name": "Locate the cup",
      "criteria": "The target cup is explicitly visible in the
                   environment observation.",
      "progress": 0.10
    },
    {
      "id": "M2",
      "name": "Pick up the cup",
      "criteria": "The agent has successfully picked up the target cup.",
      "progress": 0.25
    },
    {
      "id": "M3",
      "name": "Reach the sink with the cup",
      "criteria": "The agent is carrying the target cup and has reached
                   the sink basin.",
      "progress": 0.40
    },
    {
      "id": "M4",
      "name": "Clean the cup",
      "criteria": "The target cup has been successfully cleaned.",
      "progress": 0.65
    },
    {
      "id": "M5",
      "name": "Reach the cabinet with the clean cup",
      "criteria": "The agent is carrying the cleaned cup and has reached
                   the target cabinet.",
      "progress": 0.80
    },
    {
      "id": "M6",
      "name": "Place the clean cup in the cabinet",
      "criteria": "The cleaned target cup has been placed in the
                   specified cabinet, completing the task.",
      "progress": 1.0
    }
  ]
}
\end{lstlisting}

\subsection{Environment-Grounded Capability Evaluation Labels}
\label{app:capability_labels}

We construct capability-evaluation labels from privileged progress signals available in the two environments. These signals are used only to construct evaluation targets and are never provided to the evaluated models. For each environment, let \(P(x)\) denote the privileged progress score associated with an agent context \(x\), with larger values indicating greater environment-grounded progress toward task completion.

\paragraph{Environment-grounded progress scores.}
For \textsc{ALFWorld}, we use the environment's symbolic planner to measure
remaining distance to task completion. Let \(L_D(x)\) denote the length of an
expert plan from the restored environment state associated with context \(x\)
to a goal-satisfying state. We define
\begin{equation}
    P_{\mathrm{ALF}}(x)
    =
    1-\frac{L_D(x)}{L_D(x_0)},
\end{equation}
where \(x_0\) is the initial context of the task instance. $P_{\mathrm{ALF}}$ is formally a normalized cost-to-go-based proxy, rather than a definition of retrospective realized progress itself. We use it because \textsc{ALFWorld} consists of structured navigation and manipulation tasks in which realizing additional goal-relevant prerequisites typically reduces the remaining expert-plan distance, while losing previously satisfied conditions can increase it. We additionally exclude near-tied states when constructing pairwise labels, as described below, to avoid treating small and potentially ambiguous differences in planner distance as reliable progress distinctions. As a qualitative sanity check, we manually inspected \(100\) randomly sampled context pairs and found that the planner-induced ordering was consistent with relative realized-progress judgments in \(100\%\) of the inspected pairs.

For \textsc{ScienceWorld}, we use the benchmark's native normalized task score,
\begin{equation}
    P_{\mathrm{SW}}(x)=S_{\mathrm{env}}(x),
\end{equation}
which provides partial credit according to the task's environment-defined goals
and subgoals. Because this score directly reflects which task components have already been satisfied, it provides a more direct operationalization of realized task progress than the planner-distance proxy used in \textsc{ALFWorld}.

\paragraph{Pairwise progress labels.}
For two contexts \(x_i\) and \(x_j\) from the same task instance, we label the
context with the larger privileged progress score as having greater realized
progress. Pairs with equal scores are discarded. For \textsc{ALFWorld}, we
additionally retain a pair only when
\begin{equation}
    \left|P_{\mathrm{ALF}}(x_i)-P_{\mathrm{ALF}}(x_j)\right|>\delta,
    \qquad \delta=0.12,
\end{equation}
to exclude near-tied states for which differences in planner distance provide
an ambiguous progress ordering.

\paragraph{Anchored-localization labels.}
Each generated milestone anchor \(m_k\) is associated with a corresponding
expert-trajectory context \(\bar{x}_k\). We map this context onto the privileged
environment progress scale,
\begin{equation}
    s_k=P(\bar{x}_k),
\end{equation}
and use \(s_k\) as the environment-grounded progress level associated with
anchor \(k\). For a sampled context \(x\), the localization target is
\begin{equation}
    y_{\mathrm{loc}}(x)
    =
    \max\{k:s_k\leq P(x)\}.
\end{equation}
Thus, the target identifies the most advanced anchor whose
environment-grounded progress level has been reached by the current context.
Importantly, this construction does not require the natural-language criterion
of each milestone to be independently evaluated when constructing the target.
We use \(P_{\mathrm{ALF}}\) in \textsc{ALFWorld} and \(P_{\mathrm{SW}}\) in
\textsc{ScienceWorld}.

\textit{Pairwise comparison} and \textit{Anchored localization} therefore use the same underlying environment-grounded progress signal for evaluation: the former tests the induced ordering between contexts, while the latter tests localization relative to the fixed task-specific progress anchors.

\section{Process-Sensitive Return Derivation}
\label{app:return_derivation}

This appendix derives the process-sensitive return introduced in Section~\ref{sec:optimization_objective} and characterizes how the parameter \(\rho\) controls the temporal weighting of \textit{world potentials}. We first derive the return decomposition, then examine its boundary cases and induced trajectory preferences.

\subsection{Return Decomposition}

Recall the world-potential transition signal
\begin{equation}
r_t^{\mathrm{WP}}
=
r_t^{\mathrm{env}}
+\gamma\Phi_g(x_{t+1})
-\Phi_g(x_t),
\end{equation}
and the effective return discount
\begin{equation}
\beta=\gamma\rho,
\qquad \rho\in[0,1].
\end{equation}
For a trajectory terminating at context \(x_T\), let \(H=T-t\). The process-sensitive return from step \(t\) is
\begin{equation}
G_t^{\mathrm{WP},\rho}
=
\sum_{\ell=0}^{H-1}
\beta^\ell r_{t+\ell}^{\mathrm{WP}}.
\end{equation}
Substituting the transition signal gives
\begin{align}
G_t^{\mathrm{WP},\rho}
&=
\sum_{\ell=0}^{H-1}
\beta^\ell r_{t+\ell}^{\mathrm{env}}
+
\gamma
\sum_{\ell=0}^{H-1}
\beta^\ell
\Phi_g(x_{t+\ell+1})
-
\sum_{\ell=0}^{H-1}
\beta^\ell
\Phi_g(x_{t+\ell}).
\end{align}
Re-indexing the second potential sum yields
\begin{align}
G_t^{\mathrm{WP},\rho}
&=
\sum_{\ell=0}^{H-1}
\beta^\ell r_{t+\ell}^{\mathrm{env}}
-\Phi_g(x_t)
+
\sum_{k=1}^{H-1}
\left(
\gamma\beta^{k-1}-\beta^k
\right)
\Phi_g(x_{t+k})
+
\gamma\beta^{H-1}\Phi_g(x_T).
\end{align}
Since \(\beta=\gamma\rho\),
\begin{equation}
\gamma\beta^{k-1}-\beta^k
=
\gamma(1-\rho)\beta^{k-1}.
\end{equation}
Therefore,
\begin{align}
G_t^{\mathrm{WP},\rho}
&=
\sum_{\ell=0}^{H-1}
\beta^\ell r_{t+\ell}^{\mathrm{env}}
-\Phi_g(x_t)
\nonumber\\
&\quad+
\gamma(1-\rho)
\sum_{k=1}^{H-1}
\beta^{k-1}\Phi_g(x_{t+k})
+
\gamma\beta^{H-1}\Phi_g(x_T).
\label{eq:wp_return_decomposition_app}
\end{align}

Equation~\eqref{eq:wp_return_decomposition_app} separates the return into four components: discounted environment reward, the current-context potential, temporally weighted intermediate potentials, and the terminal-context potential. For \(0<\rho<1\), the intermediate-potential term is nonzero, allowing realized progress along the continuation to directly affect step-level credit.

\subsection{Temporal Weighting Controlled by \texorpdfstring{\(\rho\)}{rho}}

Equation~\eqref{eq:wp_return_decomposition_app} makes explicit how \(\rho\) controls the temporal weighting of \textit{world potentials}. The \(k\)-th future intermediate context receives weight
\begin{equation}
w_k(\rho)
=
\gamma(1-\rho)(\gamma\rho)^{k-1}
=
\gamma^k(1-\rho)\rho^{k-1},
\qquad
1\leq k\leq H-1,
\label{eq:app_intermediate_weight}
\end{equation}
while the terminal-context potential receives weight
\begin{equation}
w_T(\rho)
=
\gamma(\gamma\rho)^{H-1}
=
\gamma^H\rho^{H-1}.
\end{equation}
The environment reward at temporal offset \(\ell\) receives weight
\begin{equation}
u_\ell(\rho)
=
(\gamma\rho)^\ell.
\end{equation}
For two adjacent intermediate contexts,
\begin{equation}
\frac{w_{k+1}(\rho)}{w_k(\rho)}
=
\gamma\rho.
\end{equation}
Thus, \(\gamma\rho\) determines how rapidly the influence of future intermediate contexts decays with temporal distance. Smaller values of \(\rho\) concentrate credit on near-term realized progress, whereas larger values allow progress and regression farther in the future to influence the
current action.

At the same time, increasing \(\rho\) reduces the factor \(1-\rho\) multiplying intermediate potentials while increasing the relative influence of longer-range environment rewards and the terminal potential. Thus, \(\rho\) controls both the temporal range of credit and the relative emphasis on intermediate versus
terminal signals.

\paragraph{Boundary cases.}
When \(\rho=0\), we have \(\beta=0\), and therefore
\begin{equation}
G_t^{\mathrm{WP},0}
=
r_t^{\mathrm{env}}
+\gamma\Phi_g(x_{t+1})
-\Phi_g(x_t)
=
r_t^{\mathrm{WP}}.
\end{equation}
The current action is evaluated only through its immediate environment signal and one-step change in realized progress.

For \(0<\rho<1\), future intermediate contexts receive nonzero weight, so the return can distinguish continuations according to when progress is realized
and whether it is subsequently preserved.

When \(\rho=1\), the intermediate-potential coefficient vanishes:
\begin{equation}
\gamma(1-\rho)
\sum_{k=1}^{H-1}
(\gamma\rho)^{k-1}\Phi_g(x_{t+k})
=0.
\end{equation}
The potential-dependent component then reduces to
\begin{equation}
-\Phi_g(x_t)+\gamma^H\Phi_g(x_T).
\end{equation}
Thus, when the environment return and the current and terminal potentials are fixed, intermediate realized-progress differences no longer affect the return.

\paragraph{Temporal scale of an individual intermediate context.}
For a fixed temporal position $k\geq 2$, the dependence of Eq.~\eqref{eq:app_intermediate_weight} on $\rho$, up to the factor $\gamma^k$, is
\begin{equation}
f_k(\rho)
=
(1-\rho)\rho^{k-1}.
\end{equation}
Differentiating gives
\begin{equation}
\frac{d}{d\rho}
\left[
(1-\rho)\rho^{k-1}
\right]
=
\rho^{k-2}
\left[
(k-1)-k\rho
\right].
\end{equation}
Thus, the weight assigned to the $k$-th future intermediate context is maximized at
\begin{equation}
\rho_k^\star
=
\frac{k-1}{k}
=
1-\frac{1}{k}.
\label{eq:app_rho_peak}
\end{equation}
For $k=1$, $w_1(\rho)=\gamma(1-\rho)$ is instead maximized at $\rho=0$.

Equation~\eqref{eq:app_rho_peak} provides a temporal-scale interpretation: sensitivity to progress or regression occurring farther in the future requires a larger value of \(\rho\). It should not be interpreted as an optimal parameter formula for the complete learning objective, since the appropriate choice also depends on the task horizon, environment rewards, and evaluator noise.

\paragraph{Geometric-weighting interpretation.}
When $\gamma=1$, the potential component of
Eq.~\eqref{eq:wp_return_decomposition_app} becomes
\begin{equation}
-\Phi_g(x_t)
+
(1-\rho)
\sum_{k=1}^{H-1}
\rho^{k-1}\Phi_g(x_{t+k})
+
\rho^{H-1}\Phi_g(x_T).
\label{eq:app_geometric_form}
\end{equation}
The coefficients on all noncurrent contexts satisfy
\begin{equation}
(1-\rho)
\sum_{k=1}^{H-1}
\rho^{k-1}
+
\rho^{H-1}
=
1.
\end{equation}
They can therefore be viewed as a truncated geometric weighting over a future evaluation time. For the corresponding untruncated geometric distribution,
\begin{equation}
\Pr(K=k)
=
(1-\rho)\rho^{k-1},
\qquad
\mathbb{E}[K]
=
\frac{1}{1-\rho}.
\label{eq:app_geometric_mean}
\end{equation}

When $\gamma<1$, both environment and potential contributions additionally decay according to $\beta=\gamma\rho$. In this case,
\begin{equation}
H_{\mathrm{eff}}
\approx
\frac{1}{1-\gamma\rho}
\label{eq:app_effective_horizon}
\end{equation}
provides a rough temporal decay scale. Because $\gamma$ also changes the
total coefficient mass, Eq.~\eqref{eq:app_effective_horizon} should be interpreted as an influence range rather than the exact mean of a normalized distribution.

\subsection{Illustrative Trajectory Preferences}
\label{app:rho_examples}

The decomposition in the preceding subsection makes the induced trajectory preferences explicit. We illustrate these preferences with two simple examples, setting $\gamma=1$ and omitting the current-context term and environment return, which are identical across the compared trajectories.

For $\gamma=1$, the potential-dependent contribution to the return can be
written as
\begin{equation}
\widetilde G
=
(1-\rho)
\sum_{k=1}^{H-1}
\rho^{k-1}\Phi_g(x_k)
+
\rho^{H-1}\Phi_g(x_H),
\label{eq:app_example_potential_return}
\end{equation}
where $\widetilde G$ contains only the terms that can differ between the
trajectories considered below.

\subsubsection{Earlier Versus Later Realized Progress}
\label{app:early_progress_example}

Consider two trajectories that reach the same terminal potential:
\begin{align*}
\text{Trajectory A:}\quad
&
\Phi_g(x_1)=1,\quad
\Phi_g(x_2)=1,\quad
\Phi_g(x_3)=1,
\\
\text{Trajectory B:}\quad
&
\Phi_g(x_1)=0,\quad
\Phi_g(x_2)=0,\quad
\Phi_g(x_3)=1.
\end{align*}
Trajectory A realizes high potential earlier and preserves it, whereas
Trajectory B reaches the same potential only at termination.

Here $H=3$. For Trajectory A,
\begin{align}
\widetilde G_A
&=
(1-\rho)
\left[
\Phi_g(x_1)
+
\rho\Phi_g(x_2)
\right]
+
\rho^2\Phi_g(x_3)
\nonumber\\
&=
(1-\rho)(1+\rho)
+
\rho^2
\nonumber\\
&=
1.
\label{eq:app_early_A}
\end{align}
For Trajectory B,
\begin{align}
\widetilde G_B
&=
(1-\rho)
\left[
\Phi_g(x_1)
+
\rho\Phi_g(x_2)
\right]
+
\rho^2\Phi_g(x_3)
\nonumber\\
&=
\rho^2.
\label{eq:app_early_B}
\end{align}
Therefore,
\begin{align}
\Delta G_{\mathrm{early}}
&=
\widetilde G_A-\widetilde G_B
\nonumber\\
&=
1-\rho^2
\label{eq:app_early_progress}
\end{align}

Thus, for any $0<\rho<1$, the objective prefers the trajectory that realizes the same terminal progress earlier and preserves it. The strength of this preference decreases monotonically with $\rho$ and vanishes at $\rho=1$, where intermediate-context potentials no longer contribute directly to the return.

\subsubsection{Preserved Progress Versus Intermediate Regression}
\label{app:regression_example}

Now consider two trajectories with the same next-context and terminal
potentials:
\begin{align*}
\text{Trajectory A:}\quad
&
\Phi_g(x_1)=1,\quad
\Phi_g(x_2)=1,\quad
\Phi_g(x_3)=1,
\\
\text{Trajectory B:}\quad
&
\Phi_g(x_1)=1,\quad
\Phi_g(x_2)=0,\quad
\Phi_g(x_3)=1.
\end{align*}
Both trajectories make the same immediate progress and reach the same
terminal potential, but Trajectory B temporarily regresses at the second
future context.

Again, $H=3$. For Trajectory A,
\begin{align}
\widetilde G_A
&=
(1-\rho)
\left[
\Phi_g(x_1)
+
\rho\Phi_g(x_2)
\right]
+
\rho^2\Phi_g(x_3)
\nonumber\\
&=
(1-\rho)(1+\rho)
+
\rho^2
\nonumber\\
&=
1.
\label{eq:app_preserve_A}
\end{align}
For Trajectory B,
\begin{align}
\widetilde G_B
&=
(1-\rho)
\left[
\Phi_g(x_1)
+
\rho\Phi_g(x_2)
\right]
+
\rho^2\Phi_g(x_3)
\nonumber\\
&=
(1-\rho)
+
\rho^2.
\label{eq:app_preserve_B}
\end{align}
Their return difference is therefore
\begin{align}
\Delta G_{\mathrm{preserve}}
&=
\widetilde G_A-\widetilde G_B
\nonumber\\
&=
1-\left[(1-\rho)+\rho^2\right]
\nonumber\\
&=
\rho-\rho^2
\nonumber\\
&=
(1-\rho)\rho.
\label{eq:app_preserve_progress}
\end{align}

Thus, for any $0<\rho<1$, Trajectory A is preferred over Trajectory B: preserving realized progress yields a higher process-sensitive return than temporarily regressing and later recovering to the same terminal potential. At $\rho=0$, this difference is zero because a one-step objective cannot observe a regression occurring two transitions later. At $\rho=1$, it is also zero because intermediate-context potentials no longer contribute. The preference is maximized at
\begin{equation}
\rho^\star
=
\frac{1}{2}.
\end{equation}

More generally, suppose two trajectories differ in potential by  $\Delta\Phi$ only at the $k$-th future context. From Eq.~\eqref{eq:app_intermediate_weight}, the induced return difference is
\begin{equation}
\Delta G_k
=
\gamma^k
(1-\rho)
\rho^{k-1}
\Delta\Phi.
\label{eq:app_general_context_difference}
\end{equation}
For $k\geq2$, this difference is maximized with respect to $\rho$ at
\begin{equation}
\rho_k^\star
=
\frac{k-1}{k}.
\end{equation}
This illustrates that sensitivity to progress or regression farther in the future requires a larger value of $\rho$.

\subsection{Experimental Setting}
\label{app:rho_setting}

We use the same parameter setting in all policy-optimization experiments:
\begin{equation}
\gamma=0.99,
\qquad
\rho=0.95.
\end{equation}
The corresponding effective return discount is
\begin{equation}
\beta
=
\gamma\rho
=
0.9405.
\end{equation}
For intuition, Eq.~\eqref{eq:app_effective_horizon} gives the approximate
decay scale
\begin{equation}
H_{\mathrm{eff}}
\approx
\frac{1}{1-0.9405}
\approx
16.8
\end{equation}
environment steps. This setting allows substantially delayed progress and
terminal outcomes to influence earlier actions while retaining a nonzero
intermediate-potential coefficient $1-\rho=0.05$.

We use $\rho=0.95$ uniformly across tasks without per-task tuning or a
systematic sensitivity analysis. We therefore treat it as a fixed
algorithmic choice rather than claiming that it is empirically optimal or
universally preferable for long-horizon tasks.

\section{Policy Optimization Implementation Details}
\label{app:policy_optimization}

This appendix provides implementation details for the policy-optimization experiments in Section~\ref{sec:utility}. The experiments compare outcome-only GRPO with \wpm-guided optimization under a shared actor-training pipeline. The two methods use matched task sampling, rollout construction, policy models, and actor-update configurations, while differing in the learning signal and resulting credit construction.

\subsection{Environments and Task Sampling}
\label{app:task_sampling}
\textbf{Benchmark documentation.} We study two English-language interactive agent benchmarks with complementary task domains. \textsc{ALFWorld} consists of compositional household navigation and object-manipulation tasks, including object placement, examination, cleaning, heating, and cooling. \textsc{ScienceWorld} is a text-based environment grounded in an elementary-school science curriculum and contains 30 task types spanning scientific procedures and reasoning involving topics such as changes of state, measurement, electrical conductivity, biology, chemistry, and forces. Both benchmarks require multi-step interaction with simulated environments rather than static language prediction. They contain no demographic groups or human-subject data relevant to our experiments.

\noindent\textbf{Data partitions and statistics.} We use the official data partitions provided by \textsc{ALFWorld} and \textsc{ScienceWorld} and do not construct additional random train--evaluation splits. \textsc{ALFWorld} contains $3,553$ training games, $140$ valid\_seen games, and $134$ valid\_unseen games. We sample policy-training tasks from the $3,553$-game training split and evaluate on all $134$ valid\_unseen games, which contain held-out environments and task instances. \textsc{ScienceWorld} contains $7,207$ task variations across $30$ task types. Its official partitions contain $3,589$ training, $1,799$ development, and $1,819$ test variations. We sample policy-training tasks from the official training variations and evaluate on the official test variations; the development variations are not used in our policy-optimization experiments. The \textsc{ScienceWorld} splits are exposed by the environment through \texttt{get\_variations\_train()}, \texttt{get\_variations\_dev()}, and \texttt{get\_variations\_test()}.

The dataset records supplied to our trainer serve only as rollout slots and should not be interpreted as additional benchmark examples or a new data split. Interactive task instances are sampled from the corresponding official benchmark partition when the environment is initialized or reset.

\subsection{Training Configuration and Software}
\label{app:training_configuration}

\textbf{Training Configuration.} We optimize Qwen2.5-Instruct-3B, Qwen2.5-Instruct-7B, and Qwen3-4B policies. Outcome-only GRPO and \wpm-guided optimization use the same actor-training configuration. Each update samples 16 task groups with 8 policy rollouts per task, yielding 128 trajectories. Each rollout is capped at 30 environment actions, and the policy context retains the two most recent interaction steps. We use PPO-style clipped actor updates with KL regularization and the same invalid-action treatment for both methods. The learning rate is \(1\times10^{-6}\), and training runs for 100 epochs.

\noindent\textbf{Software and implementations.} Our released implementation targets Linux with NVIDIA GPUs and Python 3.12. The reproduction environment uses vLLM 0.11.0, FlashAttention 2.7.4.post1, Gymnasium 0.29.1, and Stable-Baselines3 2.6.0. We use the \textsc{ALFWorld} environment through its official Python package and the \textsc{ScienceWorld} implementation included in our released codebase. The policy checkpoints are specified through the exact Hugging Face model identifiers. World-potential evaluation and milestone generation use OpenAI-compatible vLLM endpoints serving Qwen/Qwen3-VL-32B-Thinking-FP8. The released implementation and experiment entry points are provided in our public repository. Our computational experiments were conducted using 8 NVIDIA H100 GPUs.

\subsection{Learning-Signal Construction}
\label{app:learning_signal}

\paragraph{Outcome-only GRPO.}
Let
\(
R(\tau)=10\cdot\mathbf{1}_{\mathrm{won}}
\)
denote the binary task outcome. In the policy-optimization implementation, we
use the scaled terminal environment reward
\begin{equation}
r_t^{\mathrm{env}}
=
\begin{cases}
R(\tau), & t=T-1,\\
0, & t<T-1.
\end{cases}
\end{equation}
Outcome-only GRPO constructs trajectory-level group-relative advantages from
this terminal outcome signal.

\paragraph{\wpm-guided optimization.}
World-potential supervision is produced by the frozen evaluator described in Section~\ref{sec:utility}. Milestone-ladder construction and anchored progress localization follow Appendix~\ref{app:wpm}. The process-sensitive transition signal and return follow Section~\ref{sec:optimization_objective} and Appendix~\ref{app:return_derivation}, using
\[
\gamma=0.99,
\qquad
\rho=0.95.
\]
For successful episodes, the terminal world potential is set to \(1\); otherwise, the final evaluated potential is retained.

\textit{World potentials} and process-sensitive returns are computed at the environment-action level. Step-level group-relative advantages are then constructed as described in Section~\ref{sec:optimization_objective}. Because a single environment action may contain multiple generated tokens, all valid tokens belonging to the same action share the corresponding step-level advantage. The remaining actor-update procedure is shared with outcome-only GRPO.

\subsection{Evaluation}
\label{app:policy_evaluation}

We evaluate policies using episode-level task success rate. An episode is counted as successful when its final environment output reports \texttt{info["won"] = True}. We report the percentage of successful episodes over the evaluation set. The same evaluation protocol is used for vanilla, outcome-only GRPO, and \wpm-guided policies.

\subsection{Artifact Licenses and Terms of Use}
We use publicly released benchmark environments and pretrained model checkpoints in accordance with their respective licenses and terms of use. The ALFWorld software is released under the MIT License; its official repository additionally lists TextWorld under the MIT License and Fast Downward under GPL v3.0. ScienceWorld is released under the Apache License 2.0. For the capability evaluations in Table 1, MiMo-VL-7B-RL-2508 is released under the MIT License; Qwen2.5-VL-7B-Instruct, Qwen2.5-32B-Instruct-AWQ, Qwen3-30B-A3B-Instruct-2507-FP8, Qwen3-VL-30B-A3B-Instruct, and Qwen3-VL-32B-Thinking-FP8 are released under the Apache License 2.0; and Qwen2.5-VL-72B-Instruct-AWQ is distributed under the Qwen License Agreement. For policy optimization, Qwen2.5-3B-Instruct is distributed under the Qwen Research License, which permits non-commercial research and evaluation use, while Qwen2.5-7B-Instruct and Qwen3-4B are released under the Apache License 2.0. The Qwen3-VL-32B evaluator used for world-potential estimation is also released under the Apache License 2.0. Our use of these artifacts is limited to research-oriented evaluation, inference, and policy optimization. We do not assume broader licensing rights for benchmark assets distributed separately from their software repositories, and any redistribution of third-party artifacts should retain the notices and conditions required by the corresponding upstream licenses.
\end{document}